\documentclass[conference]{IEEEtran}
\IEEEoverridecommandlockouts
\usepackage{cite}
\usepackage{amsmath,amssymb,amsfonts}
\usepackage{algorithmic}
\usepackage{graphicx}
\usepackage{textcomp}
\usepackage{xcolor}
\usepackage{blindtext}
\usepackage{multirow}  
\usepackage{balance}  
\usepackage{array} 
\usepackage{booktabs} 
\usepackage{url}  

\def\BibTeX{{\rm B\kern-.05em{\sc i\kern-.025em b}\kern-.08em
    T\kern-.1667em\lower.7ex\hbox{E}\kern-.125emX}}
\begin{document}

\title{Challenges with Extreme Class-Imbalance and Temporal Coherence: A Study on Solar Flare Data}

\author{
    \IEEEauthorblockN{Azim~Ahmadzadeh}
    \IEEEauthorblockA{\textit{dept. Computer Science} \\
        \textit{Georgia State University}\\
        Atlanta, GA, USA\\
        aahmadzadeh1@cs.gsu.edu
    }
    \and
    \IEEEauthorblockN{Maxwell~Hostetter}
    \IEEEauthorblockA{\textit{dept. Computer Science} \\
        \textit{Georgia State University}\\
        Atlanta, GA, USA \\
        mhostetter1@cs.gsu.edu
    }
    \and
    \IEEEauthorblockN{Berkay~Aydin}
    \IEEEauthorblockA{\textit{dept. Computer Science} \\
        \textit{Georgia State University}\\
        Atlanta, GA, USA\\
        baydin2@cs.gsu.edu
    }
    \and
    \IEEEauthorblockN{Manolis~K.~Georgoulis}
    \IEEEauthorblockA{\textit{RCAAM of the Academy of Athens,} \\
        Athens, Greece \\
        manolis.georgoulis@phy-astr.gsu.edu
    }
    \and
    \IEEEauthorblockN{Dustin~J.~Kempton}
    \IEEEauthorblockA{\textit{dept. Computer Science} \\
        \textit{Georgia State University}\\
        Atlanta, GA, USA\\
        dkempton1@cs.gsu.edu
    }
    \and
    \IEEEauthorblockN{Sushant~S.~Mahajan}
    \IEEEauthorblockA{\textit{dept. Physics \& Astronomy} \\
        \textit{Georgia State University}\\
        Atlanta, GA, USA \\
        mahajan@astro.gsu.edu
    }
    \and
    \IEEEauthorblockN{Rafal~A.~Angryk}
        \IEEEauthorblockA{\textit{dept. Computer Science} \\
        \textit{Georgia State University}\\
        Atlanta, GA, USA\\
        angryk@cs.gsu.edu
    }
}

\maketitle

\begin{abstract}
In analyses of rare-events, regardless of the domain of application, class-imbalance issue is intrinsic. Although the challenges are known to data experts, their explicit impact on the analytic and the decisions made based on the findings are often overlooked. This is in particular prevalent in interdisciplinary research where the theoretical aspects are sometimes overshadowed by the challenges of the application. To show-case these undesirable impacts, we conduct a series of experiments on a recently created benchmark data, named Space Weather ANalytics for Solar Flares (SWAN-SF). This is a multivariate time series dataset of magnetic parameters of active regions. As a remedy for the imbalance issue, we study the impact of data manipulation (undersampling and oversampling) and model manipulation (using class weights). Furthermore, we bring to focus the auto-correlation of time series that is inherited from the use of sliding window for monitoring flares' history. Temporal coherence, as we call this phenomenon, invalidates the randomness assumption, thus impacting all sampling practices including different cross-validation techniques. We illustrate how failing to notice this concept could give an artificial boost in the forecast performance and result in misleading findings. Throughout this study we utilized Support Vector Machine as a classifier, and True Skill Statistics as a verification metric for comparison of experiments. We conclude our work by specifying the correct practice in each case, and we hope that this study could benefit researchers in other domains where time series of rare events are of interest.
\end{abstract}

\begin{IEEEkeywords}
class imbalance, sampling, time series, flare forecast
\end{IEEEkeywords}

\section{Introduction}
    To gain valuable insights or robust predictive performance from data, we must first ensure the integrity of our data. Beyond data collection, this involves data-cleaning. It requires a thorough investigation by the experts of the domain and data scientists to produce a reliable dataset. Nonetheless, there are some challenges which are inherited from the subject under study due to unique characteristics of the data which should be identified, understood and dealt with appropriately. Class-imbalance issue is one of the main problems of this kind, which is present in many natural or other nonlinear dynamical systems. This is often due to the nature of the events, not the data collection process.\par

    \begin{figure}[t]
        \centering\includegraphics[width=\linewidth]{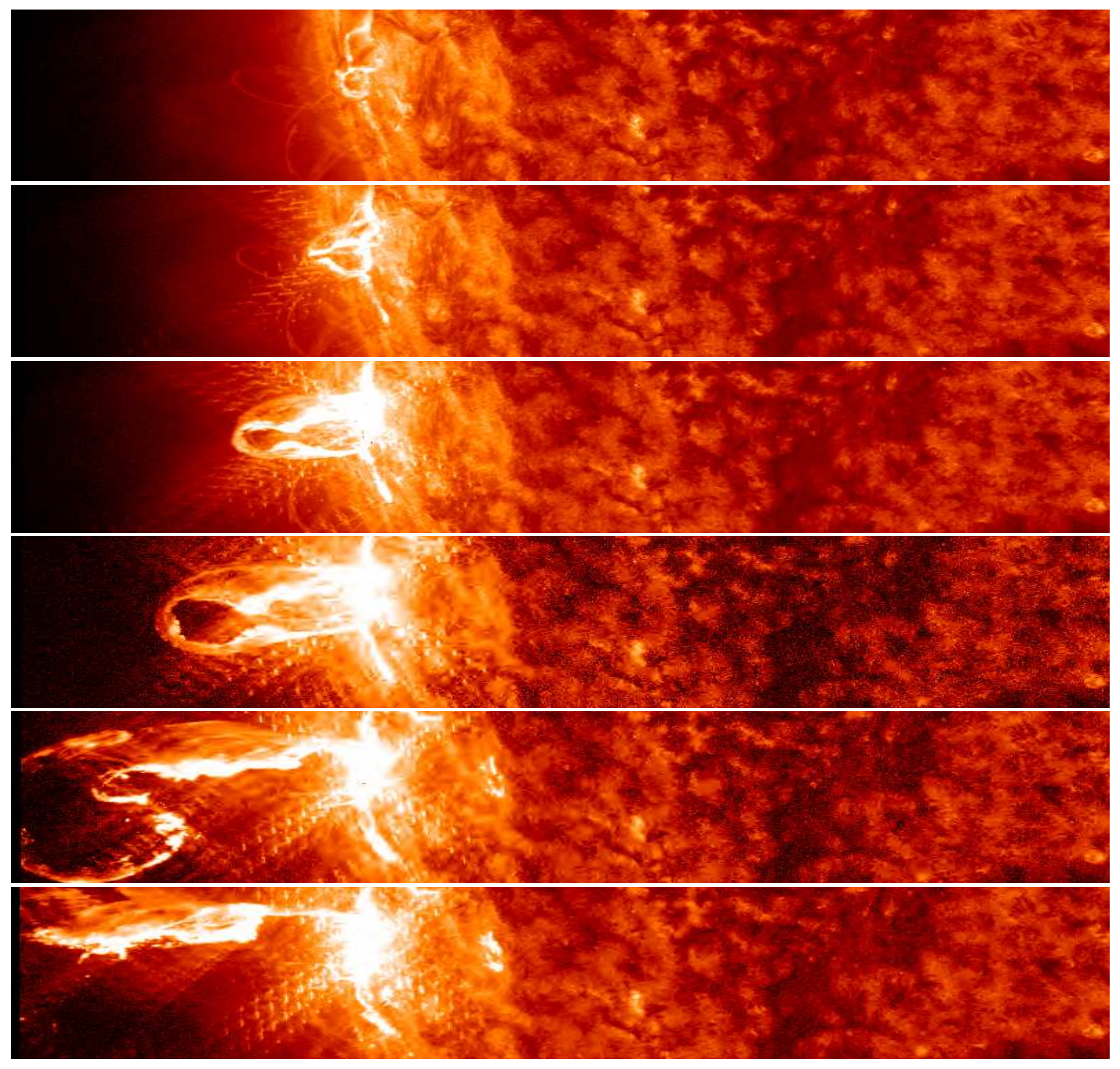}
        \caption{Snapshots of an X-class flare, peaking at 7:49 p.m. EST on Feb. 24, 2014, observed by NASA's Solar Dynamics Observatory, in the 304\AA~wavelength channel. (Images source: https://helioviewer.org/)}
        \label{fig:flareFormation}
    \end{figure}
    
    Class-imbalance is a common problem, with many potential remedies. Some of these remedies are common and well-known, but can still be misapplied. 
    This is particularly true when the primary objective is not machine learning per se but the testing and scrutiny of domain-specific theories. The complexity of the problem at hand and the absence of data experts very often underestimate the needed level of care, resulting in unrealistic and unreliable analyses, with little practical value.\par
    
    \begin{figure*}[!htb]
        \centering\includegraphics[width=0.7\textwidth]{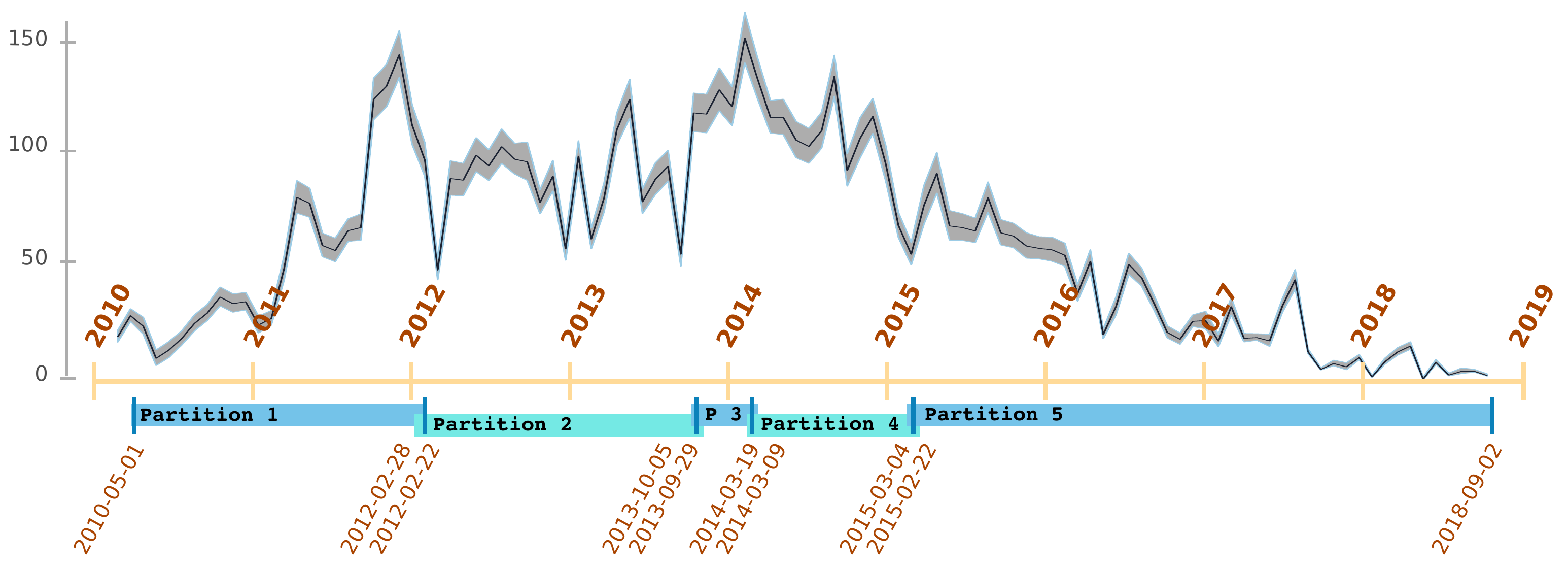}
        \caption{Time span considered for each partition of SWAN-SF dataset, the monthly report of average number of sunspots per day, and the daily variance represented with a gray ribbon (Source: WDC-SILSO, Royal Observatory of Belgium, Brussels).}
        \label{fig:dataTimeline}
    \end{figure*}
    
    In this study, we use a dataset of extracted statistical features from the time-series of solar magnetic data as an example for extreme class-imbalance. We show the impact of disregarding an interesting phenomenon called ``temporal coherence'' on classification of events in spatiotemporal datasets. We demonstrate the impact and biases of different class-imbalance remedies and discuss how they should be interpreted from the perspective of the subject under study by involving domain experts. We hope that this work raises awareness to interdisciplinary researchers and enables them to spot and tackle similar problems in their respective areas.\par
    
    All our data manipulation steps, such as preprocessing, feature extraction, dimensionality reduction, data normalization, and sampling, are made publicly available as a general-purpose python module \cite{mvtsdatatoolkt2019} that is still an active project.\par
    
\section{Common Challenges in Solar Flare Prediction}\label{sec:flarePrediction}

    Machine learning applications to solar flare forecasting are not new, but we have yet to see any significant forecast improvement resulting from these efforts. In the following sections, we present a few challenges commonly faced while tackling solar flare prediction problem from a data-driven perspective. Predicting or forecasting\footnote{In this document, we use the terms 'forecast' and 'prediction' interchangeably, as the data of this study are time series where `forecast' seems to be a better fit, while our employed statistical approach towards forecast of flares is simply a `classification' task, which is done on the history of flares. Since the classification determines the presence of flares in a fixed prediction window in future, the term `prediction' is also appropriate. This decision also allows us to avoid repetition of the word 'forecast'.} the occurrence of solar flares is a typical 21st century rare-event task. Flares are sudden and substantial enhancements of high energy electromagnetic radiation (like extreme ultra violet and X-rays) at local solar scales which pose a threat to humans and equipment in space. They are automatically detected and classified by the National Oceanic and Atmospheric Administration's (NOAA) constellation of GOES satellites based on their peak flux in soft X-ray wavelengths on a logarithmic scale as A-, B-, C-, M- and X-class solar flares. A and B-class flares are difficult to distinguish from the random variations in the Sun's background X-ray level, but C-class flares and above are detected reliably by GOES satellites. The most intense of these classes, namely M and X, are most often targeted for prediction due to their potentially adverse space-weather ramifications. An example of an X-class flare is shown in Fig.~\ref{fig:flareFormation}.\par
    
    \subsection{Extreme Class-Imbalance}\label{subsec:extremeImbalance}
        The frequency distribution of the peak X-ray fluxes of flares is nearly a perfect power law with a dynamical range spanning several orders of magnitude. A statistical analysis of NOAA's flare reports during solar cycle 23 (1995 to 2008) shows that around $50\%$ of active regions produce C-class flares, while $10\%$ produce M-class flares and less than $2\%$ produce X-class flares. Solar cycle 24 (2009 to present), from which our data discussed in Section 3 are taken \cite{aydin2019multivariate}, exhibit a much weaker major flare crop, making class-imbalance a conspicuous problem to deal with (for a review, see \cite{aschwanden201625}).\par
    
    \subsection{Point-in-time vs Time Series Forecasting}\label{subsec:pointInTime}
        Solar flares are significant precursors to other space weather phenomena. Much of the existing literature (e.g., \cite{Barnes_2016,florios2018forecasting}) is focused on using instantaneous measurements to produce a binary or probabilistic flare forecast over a preset forecast horizon. However, flares are inherently dynamical phenomena, with clear pre-flare and post-flare phases, characterized by certain evolutionary trends \cite{benz2008flare,fletcher2011observational}. Because of this, time series of aspiring flare forecasting parameters should be used, rather than isolated points in time. We believe that the sheer level of difficulty of this task was the key factor for the (over-)simplifying point-in-time assumption. However, it may be precisely this compromise that may have hampered non-incremental progress toward flare prediction. Therefore, our main goal in this study is to explore the difficulties which arise when we take into account the temporal evolution of magnetic parameters of active regions rather than looking at a single snapshot in time.\par
    
    \subsection{Non-representative Datasets}\label{subsec:nonRepresentative}
        We have access to data from many scientific instruments that map the magnetic field of the sun's photosphere. This has been monitored for 25 years \cite{tsuneta2008solar,scherrer2012helioseismic,scherrer1995solar,tritschler2016dkist,fox2018parker}, but the time span for high-quality data is still limited. Another issue for machine learning applications is that training forecasts on one part of the solar cycle is not necessarily effective for forecasting other parts of the solar cycle. This is because of continuously modulating background of magnetic activity. Some of these effects are evident in the experiments in Sec.~\ref{sec:experiments}. Therefore, there exists a problem of heterogeneous and/or non-representative data. Coupled with simpler, yet unjustified, selections of random undersampling from the majority class events, the sampled subsets of data fail to become representative of the overall flare population.\par

\section{SWAN-SF Dataset: A Multivariate Time Series Data}\label{sec:theData} 
    
    We are only just recently getting access to high-quality time-series data for use in solar flare prediction. Many previous and current projects use point-in-time measurements \cite{Barnes_2016,bobra2015solar,Bloomfield_2012, benvenuto2018hybrid, florios2018forecasting}. It is possible that a time-series approach will enable new progress on using machine learning for flare prediction.
    Here we will use a benchmark dataset, named as Space Weather ANalytics for Solar Flares (SWAN-SF), recently released by \cite{aydin2019multivariate}, and made entirely of multivariate time series, aiming to carry out an unbiased flare forecasting and hopefully set the above question to rest.\par
    
    \begin{figure}[!htb]
            \centering
            \includegraphics[width=\linewidth]{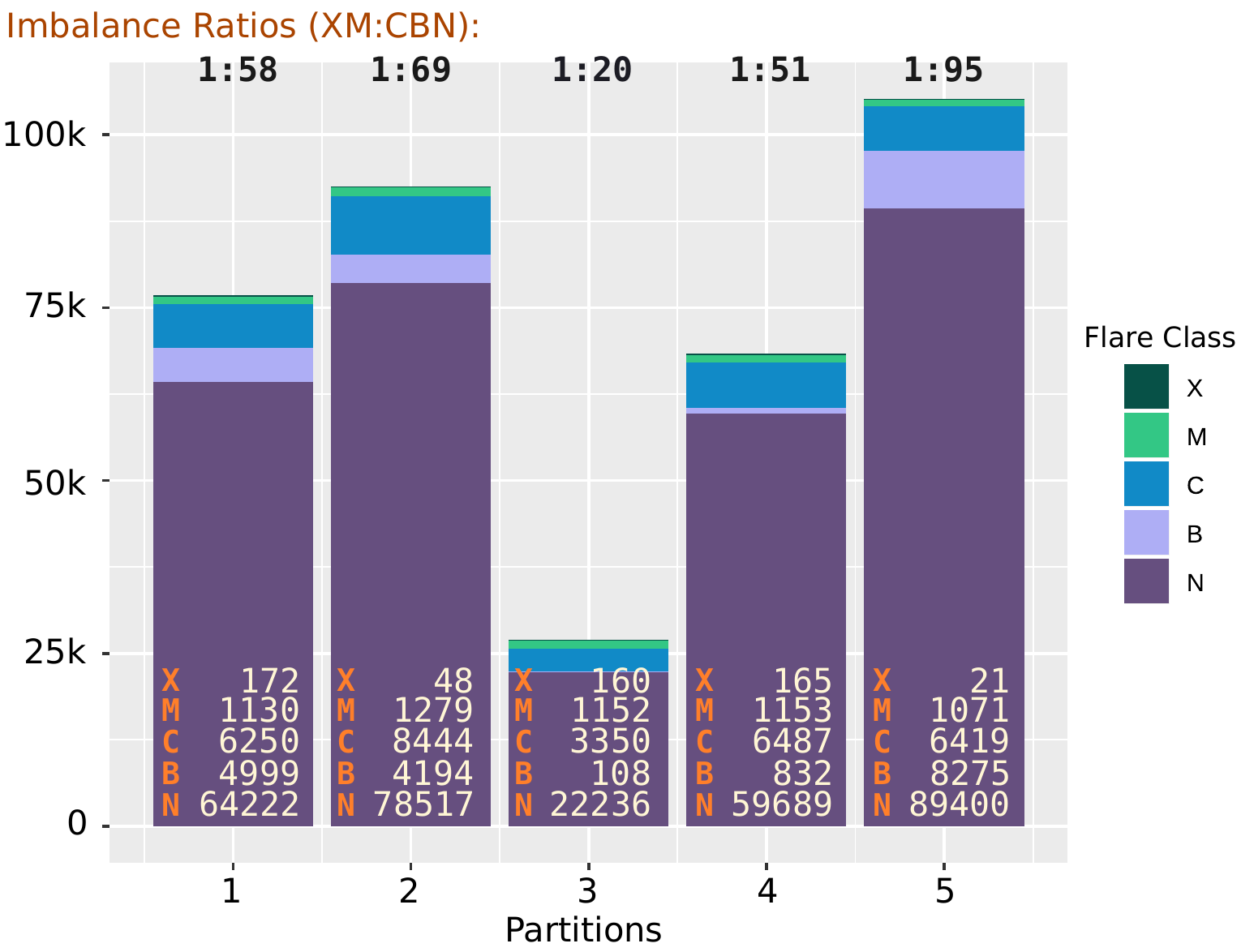}
            \caption{Counts of the five flare classes across different partitions.}
            \label{fig:classfrequencies}
    \end{figure}
        
    The SWAN-SF dataset is made of five partitions. These partitions are temporally non-overlapping and divided in such a way that each of them contains approximately an equal number of X- and M- class flares (see Fig.~\ref{fig:dataTimeline} and \ref{fig:classfrequencies}). The data points in this dataset are time series slices of physical (magnetic field) parameters extracted from the flaring and flare-quiet regions, in a sliding fashion. That is, for a particular flare with a unique id, $k$ equal-length multivariate time series are collected from a fixed period of time in the history of that flare. This period is called an \textit{observation window}, denoted by $T_{obs}$, and spans over $24$ hours. Given that $t_i$ indicates the starting point of the $i$-th slice of the multivariate time series, the $(i+1)$-th slice starts at $t_i + \tau$, where $T_{obs} = 8\tau$. This kind of sliding observation inherits the fact that very often the $k$ slices behave very similar due to their temporal and spatial closeness. In other words, the physical parameters describing the behavior of the region corresponding to a particular flare are not significantly different from one slice to the next. These similar slices, if described in the multi-dimensional feature space of our data, are located too close to each other to be considered distinct instances. And their closeness does not reflect any characteristics of those data points, except our slicing methodology. We refer to this phenomenon as \textit{temporal coherence}\footnote{We introduce this concept in the context of data manipulation, and it should not be confused with \textit{temporal coherence} in Optics or any other topic.} of data. This is a key concept to understand some of the challenges we would like to address in this study.\par
    
    \subsection{Final Forecast Dataset}\label{subsec:finalDataset}
    
        When working with time-series, we can take one of two main approaches. We can preprocess and use the data directly, with methods optimized for time-series. We can also extract a set of statistical features from the time-series and then use the extracted descriptors in our models. Our interest in the analysis of different sampling methods guides us to use the extracted features.\par

        \noindent\textbf{Statistical Features.}
        Though, as mentioned in Sec.~\ref{subsec:pointInTime}, point-in-time measurements have been proposed for flare prediction (\cite{Barnes_2016,florios2018forecasting}), there are no established theories or works on which time-series characteristics may be useful for flare forecasting. We hereby build a prediction dataset relying on the set of first four statistical moments of the time series, namely, their \textit{mean}, \textit{variance}, \textit{skewness}, and \textit{kurtosis}. To allow comparison with previous studies, we also consider a point-in-time feature, namely the \textit{last value} of each time series (from the observation window). Moreover, we also add \textit{median} to the list to compensate for the effect of outliers on \textit{mean}.\par
        
        The obtained dataset of the extracted features has a dimensionality of $144$, resulting from the computation of the $6$ above-mentioned statistics on the $24$ physical parameters of the SWAN-SF dataset. Data points of this dataset are labeled by $5$ different classes of flares, namely GOES X, M, C, B, and N. The latter represents flare-quite instances or GOES A-class events.\par
        
        This work is focused mostly on concerns with techniques common to machine learning applications, to help describe a robust methodology independent of data or model specifics. To that end, we use the \textit{last value} which is most comparable to other point-in-time studies. At the end, however, we present the contribution of other statistical features as well to show the benefit of using time series instead of point-in-time data instances. We also conduct our experiments on a binary class data by merging X and M classes into a super-class called XM, and C, B, and N classes into another super-class denoted by CBN. This simplification allows us to only focus on the challenges we mentioned before, which is the primary objective of this study.\par
        
        \noindent\textbf{Preprocessing.}
        After computing the above-mentioned features, the dataset requires a minimal preprocessing due to the presence of some missing values. Since this accounts for a very small fraction of the data (i.e., $<0.01\%$), we simply utilize linear interpolation to fill out those values. In addition, we use zero-one data transformation to normalize the data for our experiments, since otherwise the optimal hyperplanes found by SVM would be meaningless.\par

\section{Class-imbalance and Temporal Coherence}\label{sec:challenges}
    In this section, we discuss two problems that are common in machine learning and always present in data used for solar flare prediction: class-imbalance and temporal coherence. Without loss of generality, we use the SVM classifier which, like many other learners, is sensitive to these issues. Therefore, instead of analyzing the specific characteristics of each classifier, we focus on the common denominators of the well-known family of classifiers in view of imbalanced datasets.\par
    
    \subsection{Class-imbalance Problem}\label{subsec:classImbalance}
        Class-imbalance describes a case where the occurrences of one or more of the class(es) in the data are far less than that of the other class(es). When class imbalance is present, it is well-established to apply special techniques to deal with it, since generally machine learning models perform better when classes are roughly equally frequent. Here we use the terms ``minority'', or ``positive'', class to refer to the less frequent group and ``majority'', or ``negative'', class, conversely. Stronger flares (GOES X- and M- class) form our minority class, and weaker events (GOES C-, B- and A-class) belong to our majority class. From the practical point of view, we care significantly more about the minority class, since stronger flares have much bigger impact on our environment. This is why we refer to this class as `positive'. Fig.~\ref{fig:classfrequencies} illustrates the distribution of all classes in each partition.\par
        
        In classification, the aim is to optimize our objective function by minimizing the number of misclassifications. When sever class-imbalance is present, often correct classifications of the minority class are more valuable than correct classifications of the majority class. Since the density of the majority class is significantly higher than that of the minority class, many instances of the majority class may be sacrificed (i.e., misclassified) for a correct classification of an instance from the minority class. The SVM classifier in particular, searches for optimal hyper-planes to make such separations. An imbalanced dataset most likely preserves the imbalanced density of the classes even close to the decision boundaries (where the ideal class regions overlap or meet). In such a situation, a hyper-plane that is supposed to pass through the boundaries will be shifted into the region of the minority class to reduce the total number of incorrect classifications/predictions by getting all the positive classes right. This leads to higher true-negatives (i.e., correct predictions of CBN-class flares) and lower true-positives (i.e., correct predictions of XM-class flares). In other words, a model in a class-imbalance space always favors the majority class. This is of particular concern because virtually all real-life, class-imbalance problems aim to predict minority, rather than majority, events.\par
        
        The effects of class-imbalance are also seen during forecast performance evaluation. Many well-known performance metrics are significantly impacted by class-imbalance, including accuracy, precision (but not recall), and the f1-score. This is mainly because these measures ignore the number of misclassifications. For instance, on imbalance datasets, a model that classifies all instances as the negative (majority) class may result in a very high (often asymptotic to $1.0$) accuracy, while learning little or nothing about the minority class. For the particular case of class-imbalance there are defined less susceptible measures such as TSS (True Skill Statistic\footnote{This is also known as Hanssen-Kuipers Discriminant.} \cite{woodcock1976evaluation}) or the HSS (Heidke Skill Score \cite{barnes2008evaluating,mason2010testing}), with TSS being reportedly more robust for solar flare prediction \cite{Bloomfield_2012}. In this study, we use only TSS since our main objective is to show the changes in the models' performance and not to find an operation-ready model.\par
        
        \noindent\textbf{Undersampling and Oversampling.} 
        A well-known approach to handling class-imbalance is to make our training dataset balanced. This is often done by \textit{undersampling} (that is, taking out instances from the majority class) or \textit{oversampling} (that is, providing more instances to the minority class by replication). Undersampling results in using only as many negative instances (majority) as there are positive instances (minority) in the training phase, thus achieving a 1:1 balance ratio. This solution, however, comes at some cost. When undersampling, we leave out a great portion of the data during training, therefore not learning from the entire collection. To avoid the enormous data waste, a very large dataset should be available overall. When oversampling, we add replicates of the existing instances. This may cause a model to memorize the patterns and structures of minority events in the data instead of generalizing and learning about them; a bad practice that is very prone to overfitting.\par

\begin{figure}[t]
            \includegraphics[width=\linewidth]{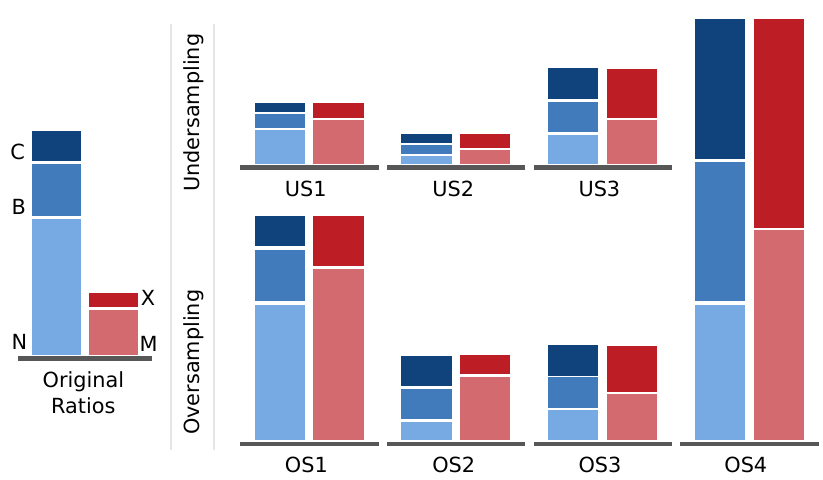}
            \caption{Visualization of different undersampling and oversampling approaches. On the left, the original proportion of the sub-classes (X, M, C, B, and N) and super-classes (XM and CBN) is depicted. Each plot on the right corresponds to one sampling approach, i.e., a single row in Table~\ref{tab:samplingExample}. This is a generic illustration and does not reflect the exact imbalance ratios in SWAN-SF.}
            \label{fig:samplingVis}
        \end{figure}

        These sampling methodologies are often used and can be effective for improving classification performance, but their applications should be specific to our data and align with the objectives of our work.
        One should be extra careful when applying them to a multi-class data, such as the flare dataset. This remains true despite the fact that we converted it to a binary class problem. In Table~\ref{tab:samplingExample}, we list a set of sampling methodologies, which are also visualized in Fig.~\ref{fig:samplingVis}. 
        \begin{table}[!htb]
            \centering
            \scriptsize
            \caption{Different undersampling and oversampling approaches.}
            \label{tab:samplingExample}
            \begin{tabular}{m{1.7cm} m{0.5cm} m{5cm}}
                \textbf{Method} & \textbf{Abbr.} & \textbf{Description}\\ 
                \firsthline
                \toprule
                Undersampling 1 & US1 & undersamples $C$, $B$, and $N$ classes while preserving their original proportions. \\ \hline
                Undersampling 2 & US2 & undersamples $C$, $B$, and $N$ classes to achieve a 1:1 balance in sub-class level, as well as the super-class level, while keeping X class unchanged.\\ \hline
                Undersampling 3 & US3 & undersamples $C$, $B$, and $N$ classes to achieve a 1:1 balance in sub-class level, while keeping $M$ class unchanged. This requires oversampling of X class.\\ \hline
                Oversampling  1 & OS1 & oversamples $X$ and $M$ classes while preserving their ratios. \\ \hline
                Oversampling  2 & OS2 & similar to OS1, except that this suppresses N-class by shrinking it down to $3|C| - (|C| + |B|)$. \\ \hline
                Oversampling  3 & OS3 & oversamples $X$ and $M$ classes to achieve a 1:1 balance in sub-class level, while keeping $C$ class unchanged. This requires undersampling of $B$ and $N$ classes.\\ \hline
                Oversampling  4 & OS4 & oversamples all fives classes to achieve a 1:1 balance in sub-class level, while keeping $N$ class unchanged. \\ \bottomrule
            \end{tabular}
        \end{table}
        As it is shown, alternative avenues exist depending on whether a 1:1 balance in the sub-class level (i.e., $|\textrm{X}|=|\textrm{M}|$ and $|\textrm{C}|=|\textrm{B}|=|\textrm{N}|$) is also required or not. Notice that this is an addition to the primary goal of our sampling which aims to achieve a balance between the super-classes (i.e., $|XM|=|CBN|$). When undersampling, for instance, if this additional balance is expected, we first need to decide which class in the minority group is considered the ``base'' class. Letting GOES X be the base class (see US2 in the table and the visualization), we must undersample from M-class flares first to balance X and M classes and then undersample from the majority (CBN) class. This yields a balanced dataset in both super-class and sub-class levels. A quick look at other approaches shows that the choice of the sampling method plays a critical role. Knowing that GOES C class represents the strongest flares in the weak-flare class (majority), it is expected that a higher fraction of C-class instances in this group results in a harder prediction problem for the model. In other words, a sampling methodology has to contort the climatology of flares to achieve the desired balance. This change affects the distribution of samples in the feature space by making the decision boundary (where GOES C and M classes overlap) denser or sparser. As an example, comparing US2 or US3 with US1 where the C-class accounts for the smallest fraction, indicates that US1 yields a slightly simpler classification problem. Having that said, US1 would still be a realistic sampling since the climatology of flares is preserved within the majority class.\par
        The above suggests that performance of different models on the same dataset is only comparable if they all employ identical sampling methodologies.\par
        
        \noindent\textbf{Misclassification Weight.} We can also modify our objective function to penalize misclassifications of the minority class more than the majority. SVM, like some other machine learning algorithms, can incorporate different weights in its objective function. For details on how this is mathematically implemented, we refer the interested readers to \cite{ben2010user}. For the experiments in this study, inspired by the class-imbalance ratio, we adjust the weights using $w_j = \frac{n}{k \cdot n_j}$, where $n$ is the total population, $n_j$ is the population in class $j$, and $k$ represents the number of classes.\par

        \subsection{Cross Validation}\label{subsec:crossValidation} 
            Cross validation is a family of statistical techniques, typically used to determine the generalization power of a model. Regardless of the employed cross validation technique ($k$-fold, leave-$p$-out, stratified or purely random), it is very often assumed that a random selection is allowed. However, in many real-world data collections, random sampling must take into account the spatiotemporal characteristics of the data, which is rooted in the temporal (or spatial) coherence of data. The temporal coherence in SWAN-SF dataset, as we discussed in Sec.~\ref{sec:theData}, prohibits this practice, since it yields an overly optimistic performance of the model's learning; a phenomenon known as `overfitting'.\par

            To avoid this mistake, we select training and testing instances from different partitions of dataset. To showcase the impact of randomly splitting the data from one partition, we conduct an experiment that is presented in Sec.~\ref{subsec:cvImpact}.

            \noindent\textbf{Validation Set.} Disregarding the temporal coherence of the data and random sub-sampling for obtaining the training and testing sets has an additional and perhaps a more important impact: when the test set is obtained by randomly splitting the data, the tested model obscures the overfitting sign, i.e., a significant difference between the training and testing performance. Therefore, the use of any sorts of sampling methodology on the test set (either for reducing the imbalance ratio, as we discussed before, or in cross validation) must be avoided at all costs if it distorts the actual distribution of the events. Cross validation is also used for optimization of models' hyper-parameters or the data-driven parameters. We wish to reiterate that for all such tasks, another independent subset of data should be used which is known as the validation set, and the test set must never be exposed to the model except for reporting the final performance of the model. For instance, to tune SVM's hyper-parameters, namely $c$ and $\gamma$ for achieving an optimal hyper-plane by the model, a validation set must be defined in order to reflect the changes that the model takes in. Based on our expectation of the acceptable performance, we may let the model improve upon the validation set's feedback. Only when we believe that the model has reached its highest performance we can use the test set to measure its robustness.\par
            
            Similar to the capital impact of different sampling methodologies on performance, as we discussed previously, sampling modification of the test set is another way of creating non-robust models with seemingly high performance.\par
        
        \subsection{Data Normalization}\label{subsec:dataNormalization}
            It is common, and depending upon our model, sometimes necessary to transform the ranges of all feature values into a unified range before training. This is called normalization and is done for a variety of reasons, most importantly for allowing the model to distinguish between different structures without taking into account the different units of measurements used for each feature. Simplicity of this concept makes researchers to sometimes overlook the variable impact of different ways of normalization. Regardless of which transformation function one may use, normalization can be done either locally and globally. Although it is a common practice to only take into account the global statistics (e.g., $min$ and $max$ of each time series in the entire dataset), for the cases similar to SWAN-SF dataset, this must be put to the test. Looking at our dataset, among the five partitions spanning over a period of ten years, the changes in solar magnetic activity from minimum to maximum during the solar cycle makes the statistical characteristics of different partitions significantly different. In such cases, a local normalization would impact some features negatively, or even generate some out-of-range values. We investigate this via another experiment in Sec.~\ref{subsec:normalizationImpact}, by showing the significantly different performance levels using different statistics.\par
    
        \subsection{Hyper-parameter Tuning}\label{subsec:hyperparameterTuning}
            When performing hyper-parameter tuning, we should take temporal coherence into account. Any supervised learning model requires optimization of its hyper-parameters in a data-driven manner. Since the discovered hyper-parameters should remain optimal over the entire dataset (including the upcoming data points for a deployed model) the training and validation sets, as well as the test set, must each be representative of the entire dataset. In our dataset, because of the temporal coherence, random sampling does not produce such subsets. Hence, tuning process remains confined to the partitions. Although the solution we proposed for cross validation using NOAA AR Number, may also be used to tackle this problem, it is very likely that a model highly optimized on one partition simply may be not performing well globally. Therefore, we believe that this is a problem yet to be investigated more thoroughly since at this point it is very clear to us that flare forecast problem has a dynamic and periodic behavior, for which ensemble models may be more appropriate.\par

\section{Experiments, Results, and Interpretations}\label{sec:experiments}
    
    In this section, we present a series of experiments, each designed to address one of the problems we have previously discussed. We also elaborate on the interpretation of these experiments in regards to our overarching flare forecasting task. \par

    Our objective is not to train the next best flare forecasting model, but to compare models that were trained differently, to help establish data driven practices recommended for solar flare forecasting. Therefore, although any changes on the data (e.g., using different normalization, data split, or sampling techniques) require re-tuning of the hyper-parameters, without loss of generality we rely on our pre-tuned hyper-parameters for SVM: $c=1000$, $\gamma=0.01$, with a Radial Based Function (RBF) kernel.\par

    We use TSS to evaluate the results of each experiment only to show the changes different treatments cause. A high TSS does not necessarily indicate a robust forecast and is often only presented with an accompanying HSS. We do not present the HSS because we evaluate our results on all different partitions of our dataset which do not have equal levels of class-imbalance. HSS is susceptible to changes in class-imbalance and is not ideal for comparison between trials in our case. \par

    \subsection{Baseline}\label{subsec:baseline}
        We establish a baseline performance by training the model without applying any class-imbalance remedies. \par
        
        \noindent\textbf{Experiment Z: Baseline.} This experiment is as simple as training SVM on all instances of one partition and testing the model on another partition. We try this on all possible partition pairs, resulting in $20$ different trials, to illustrate how the difficulty of the prediction task varies as the partitions are chosen from different parts of the solar cycle. The results are visualized in Fig.~\ref{fig:allRemedies}, along with the impact of discussed class-imbalance remedies that we further discuss in the following sections.\par
        
    \subsection{Tackling Class-imbalance Issue}\label{subsec:remedy-undersampling}
        In Sec.~\ref{subsec:classImbalance}, we discussed three different approaches towards tackling the class-imbalance problem. To show the impact of each solution, we carry out some experiments and discuss the results below.\par
        
        In the following experiments, we train and test SVM on all $20$ permutations of the partition pairs independently. In each round, the model learns from instances in the training partition and is then tested against the (different) testing partition. To measure the confidence of a model's performance when a certain sampling method is employed, we repeat the experiment $10$ times and report the variance and mean value of TSS.\par

\begin{figure}[!htb]
            \centering
            \includegraphics[width=\linewidth]{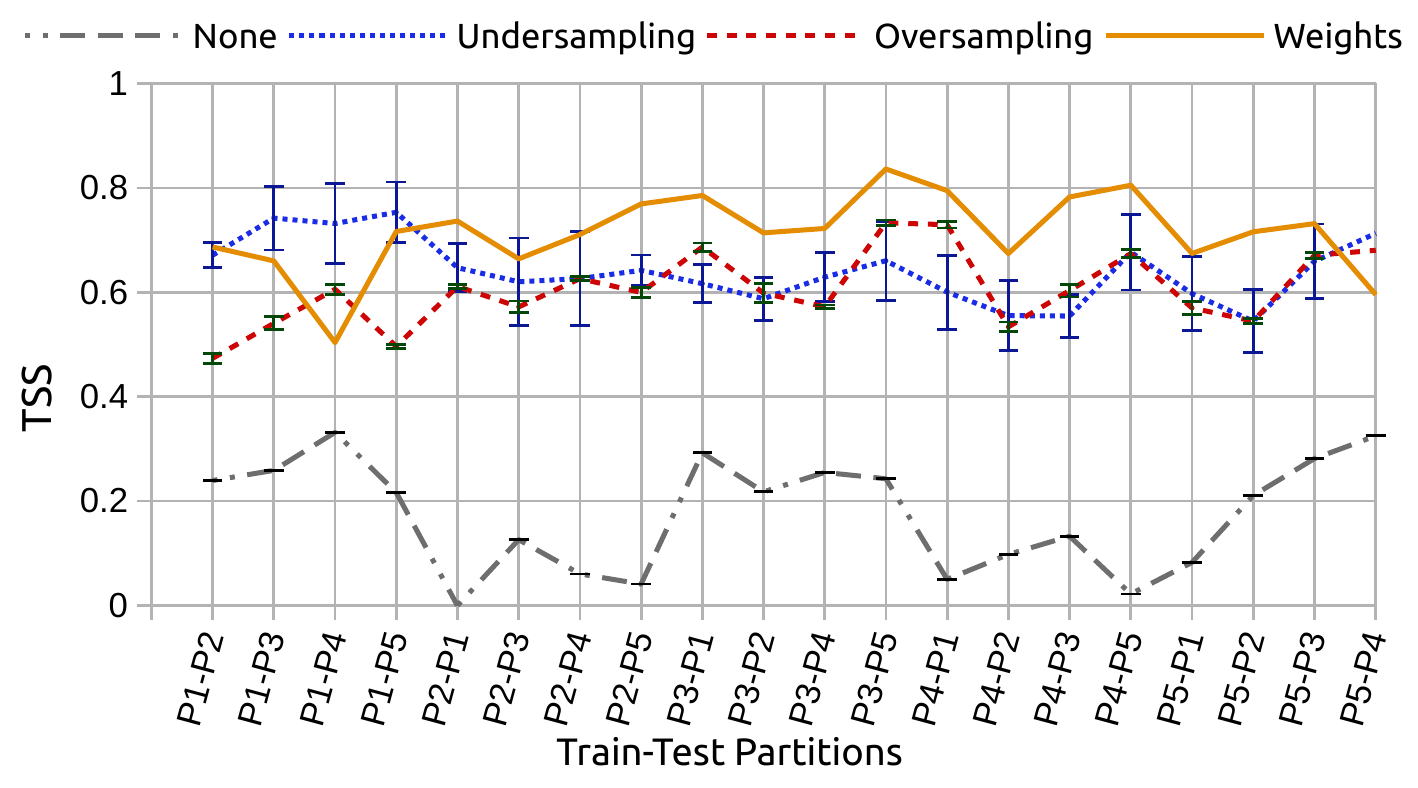}
            \caption{[Experiments Z, A, B, and C] Average TSS of SVM trained and tested on all possible permutations of partition pairs, using three different remedies for the class-imbalance issue (undersampling, oversampling, and misclassification weights), compared to a baseline performance where no remedy is employed. In all cases, global normalization is used.}
            \label{fig:allRemedies}
        \end{figure}

        \noindent\textbf{Experiment A: Undersampling.} During the training phase, we apply Undersampling 2 (US2 from Table~\ref{tab:samplingExample}) which is an X-class based undersampling. This enforces a $1:1$ balance, not only in the super-class level (i.e., $|\textrm{XM}|=|\textrm{CBN}|$) but also in the sub-class level (i.e., $|\textrm{X}|=|\textrm{M}|$ and $|\textrm{C}|=|\textrm{B}|=|\textrm{N}|$). The trained model is then tested against all other partitions one by one to examine the robustness of the model. The undersampling step is only taken in the training partition, as undersampling of the test partition distorts reality and would not reflect the true model's performance. The consistent and significant impact of this remedy is evident in Fig.~\ref{fig:allRemedies}.\par

        \noindent\textbf{Experiment B: Oversampling.} Here, we use Oversampling 3 (OS3 of Table~\ref{tab:samplingExample}) and perform an experiment similar to experiment A. Again, no over- or undersampling takes place in the testing set. Comparing the results of oversampling with Undersampling 2, in Fig.~\ref{fig:allRemedies}, shows a close correspondence between the two models in terms of their mean TSS values; typically, differences are within applicable uncertainties.\par

        \noindent\textbf{Experiment C: Mis-classification Weights.} Here, we apply weights as discussed in Sec.~\ref{subsec:classImbalance}. We use the imbalance ratio of the super-classes as the weights. For instance when working with \textit{Partition} 3, since the minority-to-majority ratio is $1:20$, we set $w_{\textrm{XM}}=20$ and $w_{\textrm{CBN}}=1$. As shown in Fig.~\ref{fig:allRemedies}, this solution outperforms both undersampling and oversampling approaches in terms of their TSS. It is worth pointing out that employing misclassification weights has the advantage of a data-driven tunability that may be better suited than over- and undersampling to achieve more robust forecast models.\par

    \subsection{Impact of Cross Validation}\label{subsec:cvImpact}
        In this experiment we demonstrate the importance of temporal coherence and the potential pitfalls of random sampling that we discussed in Sec.~\ref{subsec:crossValidation}. \par
        
        \noindent\textbf{Experiment D: Data Splits.} To demonstrate the overfitting that occurs when we do not account for the temporal coherence, we train and test SVM on instances randomly chosen from the same partition. Technically, this is a $k$-fold cross validation using a random sub-sampling method with $k=10$. The results are then juxtaposed with those obtained by training SVM on one partition and testing it on another. We equipped SVM in both scenarios with misclassification weights (the same as in Experiment C), to eliminate the need for an additional sampling layer. Therefore, the only determining factor is whether the instances are sampled from the same partition or not. Let it be clear that sampling from a single partition does not mean any overlapping between the training and testing sets.\par 
        
\begin{figure}[!htb]
            \centering
            \includegraphics[width=\linewidth]{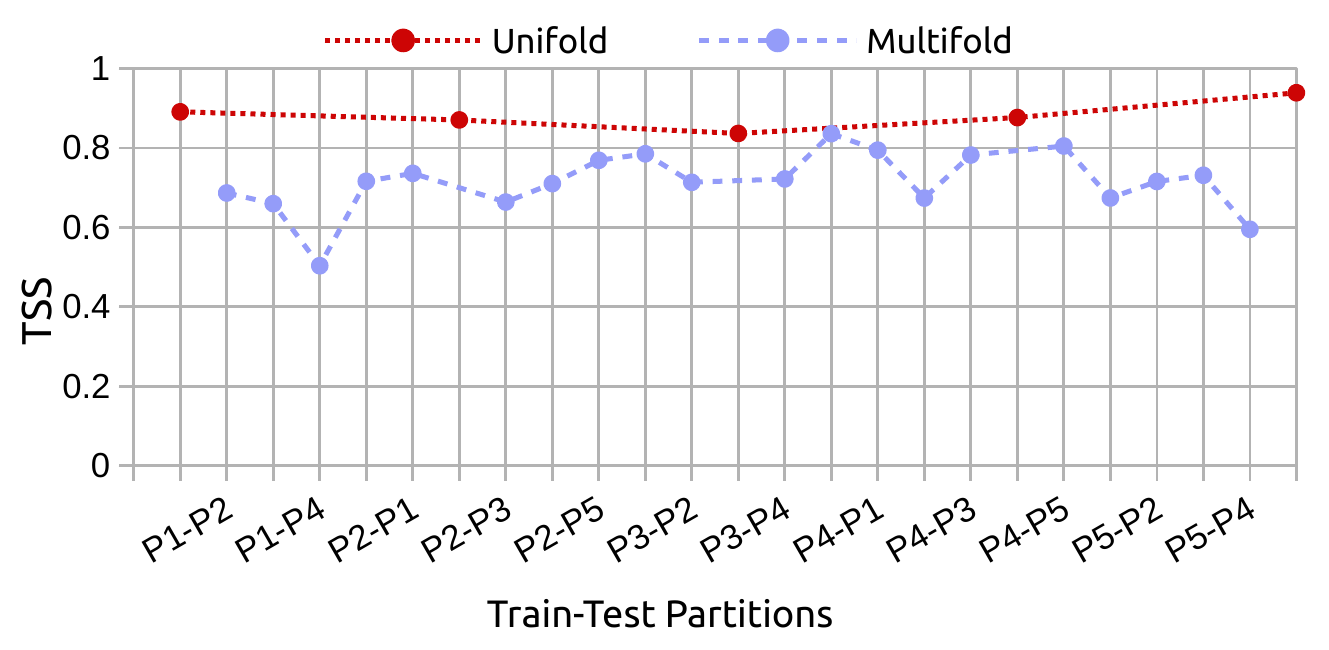}
            \caption{[Experiment D] Average TSS of SVM performance when trained and tested either using a 10-fold cross validation sub-sampling on a single partition (unifold) or assigning different partitions for training and testing (multifold). In all cases, global normalization is used and the SVM has been equipped with misclassification weights.}
            \label{fig:unifold-multifold}
        \end{figure}

        The results are summarized in Fig.~\ref{fig:unifold-multifold}. When SVM is trained and tested on a single partition (denoted as `Unifold' in the figure), performance is boosted very significantly with TSS $\in (0.84, 0.94)$ with an average TSS$\simeq 0.88$. Training and testing on different partitions (denoted as `Multifold' in the figure) yields TSS $\in (0.50, 0.84)$ with an average TSS$\simeq 0.71$. This remarkable difference should not be viewed as evidence of the robustness of the model but rather points to memorization and hence overfitting, caused when a forecast model is trained and tested on a temporally coherent dataset. It is the lower performance when the model is trained and tested on different partitions that better encapsulates its true robustness.\par
        
    \subsection{Impact of Normalization}\label{subsec:normalizationImpact}
        To demonstrate the impact of the different normalization techniques discussed in Sec.~\ref{subsec:dataNormalization}, we conducted the following experiment. \par
        
        \noindent\textbf{Experiment E: Normalization.} To transform the feature space to a normalized space (using a zero--one normalization method) we apply global normalization on a pair of partitions by using global extrema ($min$ and $max$). We then train SVM on one partition and test on the other. In another attempt, we apply normalization on each partition separately, using the local extrema of the corresponding partition. In order the minimize the impact of other factors on our experiment, we avoid employing any remedies for handling the class-imbalance issue.\par
        
        \begin{figure}[!htb]
            \centering
            \includegraphics[width=\linewidth]{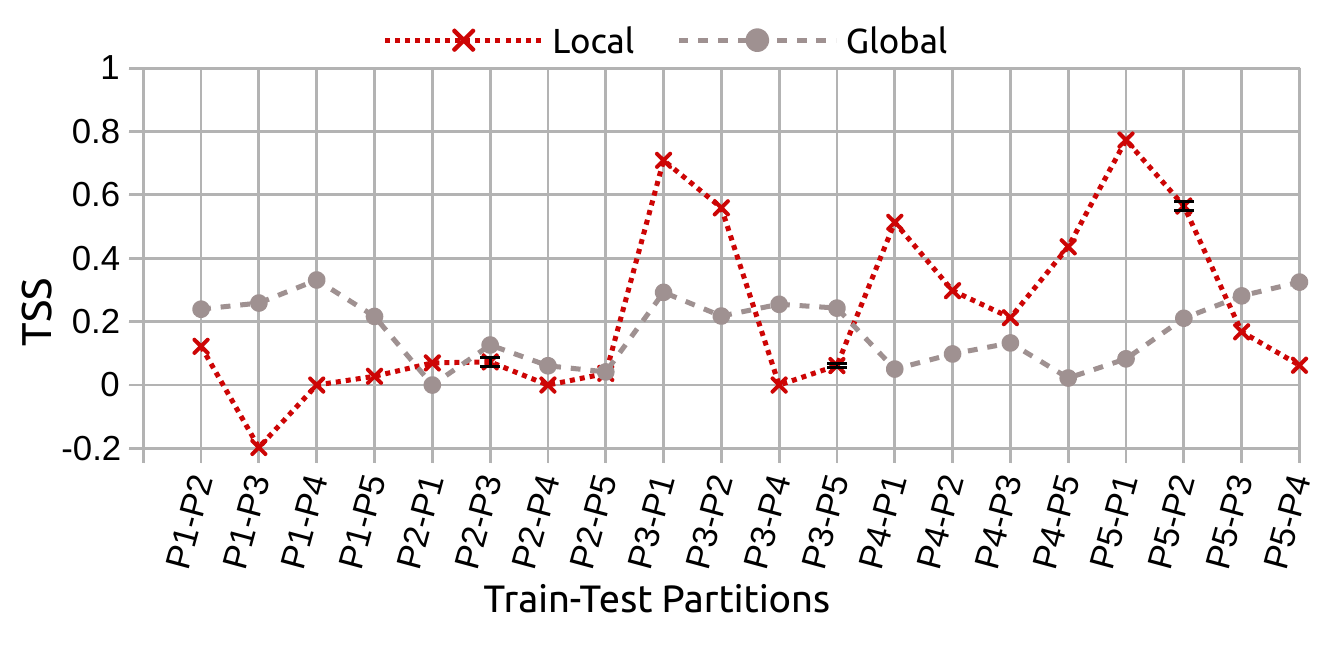}
            \caption{[Experiment E] Average TSS of SVM performance impacted by two different data normalization approaches; global and local. In all $20$ trials shown, no special remedy to the class-imbalance problem has been employed.}
            \label{fig:normalization-none}
        \end{figure}
    
        In Fig.~\ref{fig:normalization-none}, we can see dramatic differences between these two applications of normalization. This is because of the continuously modulating magnetic activity during the 11-year solar cycle.
        This makes the normalization inconsistent between the training and the testing partitions in case of local normalization. Careful observation reveals that, when local normalization is utilized, it seems training on the second half of the solar cycle (\textit{Partitions} 3, 4, and 5) builds models which perform better when tested on the instances from the first half of the cycle (\textit{Partitions} 1 and 2). Conversely, when models are trained on the first two partitions and tested on the rest, their TSS drops significantly. Both of these observations strongly agree on the noticeable differences in data collected from different windows of time. On the other hand, global normalization leads to more robust models, as the changes in TSS values are relatively lower, compared to those in local normalization. This is achieved at the cost of a sizable decrease in the overall performance. While additional investigations of this effect are clearly warranted, one gathers that performance may be optimized when training and testing take place on similar levels of solar activity, when local and global normalization converge on similar normalized values. This, however, is hardly tenable in operational settings, when forecast models are standing by to perform on yet unknown day-by-day solar activity levels.\par

    \subsection{Oversampling Impact}\label{subsec:oversamplingImpact}
        In Sec.~\ref{subsec:classImbalance}, we showed that there are multiple variants of oversampling and undersampling approaches. We also presented how this affects flare distributions in \textit{Partition} 3 as an example. Below we test how different oversampling techniques impact TSS values across different partitions.\par
        
\begin{figure}[!hbt]
            \centering
            \includegraphics[width=\linewidth]{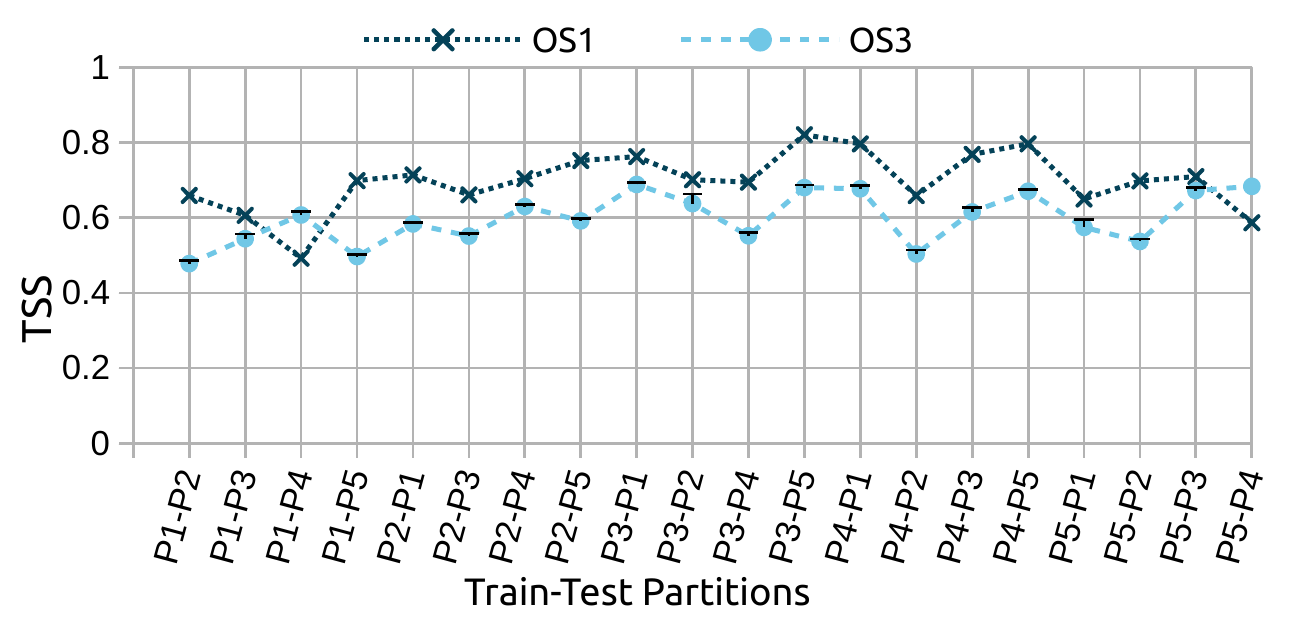}
            \caption{[Experiment F] Average TSS of SVM performance impacted by two different oversampling methods from Table~\ref{tab:samplingExample}: Oversampling 1 (OS1), where the climatology of sub-classes are preserved, and Oversampling 3 (OS3), where the sub-classes are forced to reach a $1:1$ balance ratio by considering C-class to be the base. In all cases, global normalization is used.}
            \label{fig:differentOversamplings}
        \end{figure}

        \noindent\textbf{Experiment F: Oversampling with or without Sub-Class Balance.} We use Oversampling 1 and 3 (from Table~\ref{tab:samplingExample}) in the training phase to remedy the class-imbalance problem and then we test the trained model against all other partitions. We chose OS1 and OS3 since their differences make them an interesting pair; OS1, compared to OS3, replicates M-class instances with a significantly larger factor than it does with X-class instances, allowing a large number of `easier' instances in terms of classification. Therefore, it is naturally expected that OS1 results in an `easier' dataset, hence a higher classification performance. Here we use global normalization without meaning to imply that it performs necessarily better. Our results are shown in Fig.~\ref{fig:differentOversamplings}. For them, one sees a relatively similar, consistent performance, although the climatology-preserving Oversampling 1 gives a statistically higher performance. This said, it becomes clear that different oversampling methods give non-identical performances. Therefore, comparison of any two forecasting models on similar datasets will be fair only if the employed sampling methodologies are identical.\par
        
    \subsection{Using Other Time Series Features}\label{subsec:otherFeatures}
        As mentioned in Sec.~\ref{subsec:finalDataset}, all previous experiments involved using only the \textit{last value} feature. In this experiment, we present one of the benefits of using time-series as opposed to point-in-time measurements. \par

\begin{figure}[t]
            \centering
            \includegraphics[width=\linewidth]{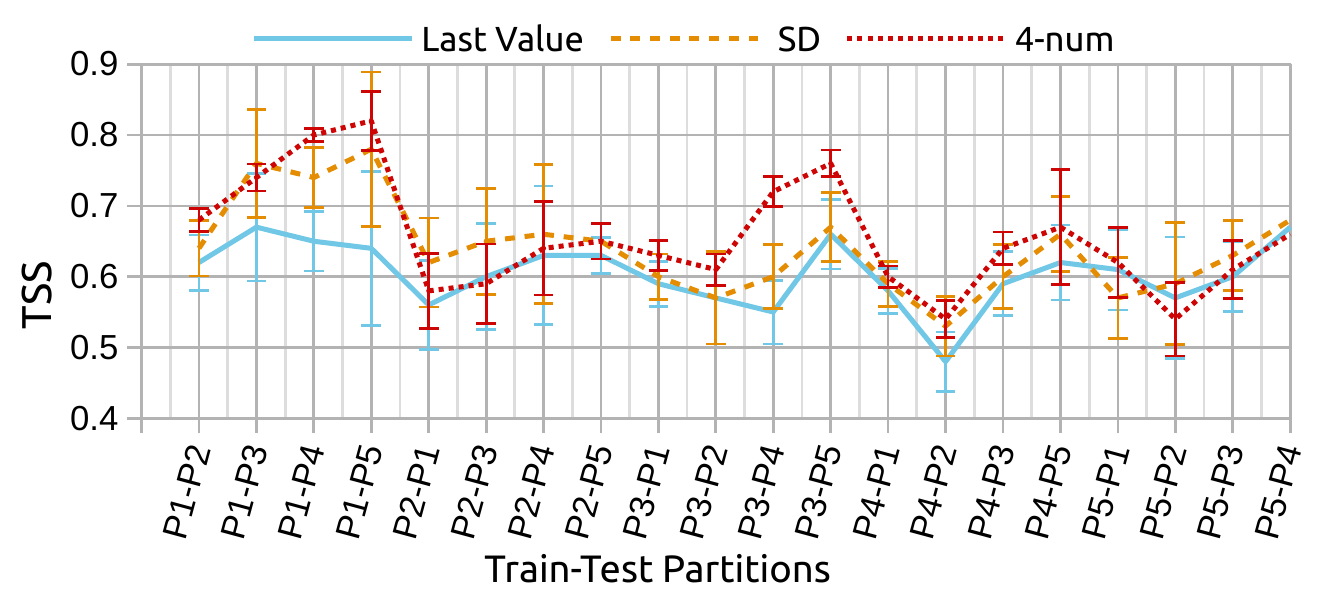}
            \caption{[Experiment G] Average TSS of SVM performance on 3 different feature spaces: 1) \textit{last value}, 2) \textit{standard deviation}, and 3) four-number summary of time series, namely \textit{median}, \textit{standard deviation}, \textit{skewness}, and \textit{kurtosis}. Undersampling 2 (i.e., US2 from Table~\ref{tab:samplingExample}) is used to remedy the class-imbalance issue. In all cases, global normalization is used.}
            \label{fig:differentFeatures}
        \end{figure}

        \noindent\textbf{Experiment G: SVM with Other Statistical Features.} During training in this experiment, we apply Undersampling 2 from Table~\ref{tab:samplingExample}. We train and test SVM on all partition pairs using (i) \textit{last value}, (ii) \textit{standard deviation}, and (iii) \textit{median}, \textit{standard deviation}, \textit{skewness}, \textit{kurtosis}. As illustrated in Fig.~\ref{fig:differentFeatures}, \textit{standard deviation} results in statistically better performance than \textit{last value}, and using the four-number summary seems to outperform \textit{standard deviation}. This is a very good indication that different characteristics of time series carry some important pieces of information that may significantly improve reliability of a forecast model.\par
            
        We leave the investigation for determining the optimal number of statistical features or selection of which statistical features are most effective for improving flare forecast performance for future studies. \par

\section{Summary, Conclusions, and Future Work}\label{sec:summary}
    We used SWAN-SF benchmark dataset as a case study to highlight some of the challenges in working with imbalanced datasets, which are very often overlooked by scientists of the domain. We also addressed an interesting characteristic of some datasets, that we called \textit{temporal coherence}, inherited from the spatial and temporal dimensions of the data. Using several different experiments, we showcased some pitfalls and overlooked consequences of disregarding those peculiarities, and we discussed the impact of different remedies in the context of flare forecast problem.\par

    There are still many interesting cases left to be discussed that we plan to include in our future studies. In several experiments, for instance, we noticed that despite the improvement in models' performances in terms of TSS, other measures such as HSS showed a moderate deterioration in the models. A new measure that reflects both these skill scores appears to be necessary to avoid many misleading interpretations. In spite of many studies, it is still an unsolved problem. Another avenue for further investigation is to incorporate NOAA AR Numbers in the sampling phase so that the temporal coherence of data that confines normalization and hyper-parameter tuning tasks to only subsets of data, can be bridged over.\par

    We hope that this work raises awareness not only to the scientists interested in flare forecast problem, but all interdisciplinary researchers who might be dealing with imbalanced and temporally coherent datasets, and enables them to spot and tackle similar problems in their respective areas.\par

\section*{Acknowledgment}
    This work was supported in part by two NASA Grant Awards [No. NNX11AM13A, and No. NNX15AF39G], and one NSF Grant Award [No. AC1443061]. The NSF Grant Award has been supported by funding from the Division of Advanced Cyberinfrastructure within the Directorate for Computer and Information Science and Engineering, the Division of Astronomical Sciences within the Directorate for Mathematical and Physical Sciences, and the Division of Atmospheric and Geospace Sciences within the Directorate for Geosciences. Also, we would like to mention that all images used in this work are courtesy of NASA/SDO and the AIA, EVE, and HMI science teams.\par

\balance

\bibliographystyle{IEEEtran}
\bibliography{RareEvent_Main}

\begin{thebibliography}{10}
\providecommand{\url}[1]{#1}
\csname url@samestyle\endcsname
\providecommand{\newblock}{\relax}
\providecommand{\bibinfo}[2]{#2}
\providecommand{\BIBentrySTDinterwordspacing}{\spaceskip=0pt\relax}
\providecommand{\BIBentryALTinterwordstretchfactor}{4}
\providecommand{\BIBentryALTinterwordspacing}{\spaceskip=\fontdimen2\font plus
\BIBentryALTinterwordstretchfactor\fontdimen3\font minus
  \fontdimen4\font\relax}
\providecommand{\BIBforeignlanguage}[2]{{%
\expandafter\ifx\csname l@#1\endcsname\relax
\typeout{** WARNING: IEEEtran.bst: No hyphenation pattern has been}%
\typeout{** loaded for the language `#1'. Using the pattern for}%
\typeout{** the default language instead.}%
\else
\language=\csname l@#1\endcsname
\fi
#2}}
\providecommand{\BIBdecl}{\relax}
\BIBdecl

\bibitem{mvtsdatatoolkt2019}
\BIBentryALTinterwordspacing
A.~Ahmadzadeh and K.~Sinha, ``A multivariate time series data toolkit,'' Nov.
  2019. [Online]. Available:
  \url{https://bitbucket.org/gsudmlab/mvtsdata_toolkit/}
\BIBentrySTDinterwordspacing

\bibitem{aydin2019multivariate}
B.~Aydin \emph{et~al.}, ``Multivariate time series dataset for space weather
  data analytics (manuscript submitted for publication),'' \emph{Scientific
  Data}, 2019.

\bibitem{aschwanden201625}
\BIBentryALTinterwordspacing
M.~J. Aschwanden \emph{et~al.}, ``25 years of self-organized criticality: solar
  and astrophysics,'' \emph{Space Science Reviews}, vol. 198, no. 1-4, pp.
  47--166, 2016. [Online]. Available:
  \url{https://link.springer.com/article/10.1007\%2Fs11214-014-0054-6}
\BIBentrySTDinterwordspacing

\bibitem{Barnes_2016}
\BIBentryALTinterwordspacing
G.~Barnes \emph{et~al.}, ``A comparison of flare forecasting methods. i.
  results from the \textquotedblleft all-clear \textquotedblright workshop,''
  \emph{The Astrophysical Journal}, vol. 829, no.~2, p.~89, Sep 2016. [Online].
  Available:
  \url{https://iopscience.iop.org/article/10.3847/0004-637X/829/2/89}
\BIBentrySTDinterwordspacing

\bibitem{florios2018forecasting}
\BIBentryALTinterwordspacing
K.~Florios, I.~Kontogiannis, S.-H. Park, J.~A. Guerra, F.~Benvenuto, D.~S.
  Bloomfield, and M.~K. Georgoulis, ``Forecasting solar flares using
  magnetogram-based predictors and machine learning,'' \emph{Solar Physics},
  vol. 293, no.~2, p.~28, Jan 2018. [Online]. Available:
  \url{https://doi.org/10.1007/s11207-018-1250-4}
\BIBentrySTDinterwordspacing

\bibitem{benz2008flare}
\BIBentryALTinterwordspacing
A.~Benz, ``Flare observations,'' \emph{Living Reviews in Solar Physics},
  vol.~5, 02 2008. [Online]. Available: \url{doi.oeg/10.12942/lrsp-2008-1}
\BIBentrySTDinterwordspacing

\bibitem{fletcher2011observational}
\BIBentryALTinterwordspacing
L.~Fletcher, B.~R. Dennis, H.~S. Hudson, S.~Krucker, K.~Phillips, A.~Veronig,
  M.~Battaglia, L.~Bone, A.~Caspi, Q.~Chen, P.~Gallagher, P.~T. Grigis, H.~Ji,
  W.~Liu, R.~O. Milligan, and M.~Temmer, ``An observational overview of solar
  flares,'' \emph{Space Science Reviews}, vol. 159, no.~1, p.~19, Aug 2011.
  [Online]. Available: \url{https://doi.org/10.1007/s11214-010-9701-8}
\BIBentrySTDinterwordspacing

\bibitem{tsuneta2008solar}
\BIBentryALTinterwordspacing
S.~Tsuneta, K.~Ichimoto, Y.~Katsukawa, S.~Nagata, M.~Otsubo, T.~Shimizu,
  Y.~Suematsu, M.~Nakagiri, M.~Noguchi, T.~Tarbell, A.~Title, R.~Shine,
  W.~Rosenberg, C.~Hoffmann, B.~Jurcevich, G.~Kushner, M.~Levay, B.~Lites,
  D.~Elmore, T.~Matsushita, N.~Kawaguchi, H.~Saito, I.~Mikami, L.~D. Hill, and
  J.~K. Owens, ``The solar optical telescope for the hinode mission:
  An overview,'' \emph{Solar Physics}, vol. 249, no.~2, pp. 167--196, Jun
  2008. [Online]. Available: \url{https://doi.org/10.1007/s11207-008-9174-z}
\BIBentrySTDinterwordspacing

\bibitem{scherrer2012helioseismic}
\BIBentryALTinterwordspacing
P.~H. Scherrer, J.~Schou, R.~I. Bush, A.~G. Kosovichev, R.~S. Bogart, J.~T.
  Hoeksema, Y.~Liu, T.~L. Duvall, J.~Zhao, A.~M. Title, C.~J. Schrijver, T.~D.
  Tarbell, and S.~Tomczyk, ``The helioseismic and magnetic imager (hmi)
  investigation for the solar dynamics observatory (sdo),'' \emph{Solar
  Physics}, vol. 275, no.~1, pp. 207--227, Jan 2012. [Online]. Available:
  \url{https://doi.org/10.1007/s11207-011-9834-2}
\BIBentrySTDinterwordspacing

\bibitem{scherrer1995solar}
\BIBentryALTinterwordspacing
P.~H. Scherrer, R.~S. Bogart, R.~I. Bush, J.~T. Hoeksema, A.~G. Kosovichev,
  J.~Schou, W.~Rosenberg, L.~Springer, T.~D. Tarbell, A.~Title, C.~J. Wolfson,
  and I.~Zayer, \emph{The Solar Oscillations Investigation --- Michelson
  Doppler Imager}.\hskip 1em plus 0.5em minus 0.4em\relax Dordrecht: Springer
  Netherlands, 1995, pp. 129--188. [Online]. Available:
  \url{https://doi.org/10.1007/978-94-009-0191-9_5}
\BIBentrySTDinterwordspacing

\bibitem{tritschler2016dkist}
\BIBentryALTinterwordspacing
A.~Tritschler, T.~R. Rimmele, S.~Berukoff, R.~Casini, J.~R. Kuhn, H.~Lin, M.~P.
  Rast, J.~P. McMullin, W.~Schmidt, F.~Woger, and D.~Team, ``Daniel k. inouye
  solar telescope: High-resolution observing of the dynamic sun,''
  \emph{Astronomische Nachrichten}, vol. 337, no.~10, pp. 1064--1069, 11 2016.
  [Online]. Available:
  \url{https://onlinelibrary.wiley.com/doi/abs/10.1002/asna.201612434}
\BIBentrySTDinterwordspacing

\bibitem{fox2018parker}
N.~{Fox}, ``{Parker Solar Probe: A NASA Mission to Touch the Sun},'' in
  \emph{EGU General Assembly Conference Abstracts}, vol.~20, Apr 2018, p.
  10345.

\bibitem{bobra2015solar}
\BIBentryALTinterwordspacing
M.~G. Bobra and S.~Couvidat, ``Solar flare prediction using sdo/hmi vector
  magnetic field data with a machine-learning algorithm,'' \emph{The
  Astrophysical Journal}, vol. 798, no.~2, p. 135, 2015. [Online]. Available:
  \url{doi.org/10.1088/0004-637X/798/2/135}
\BIBentrySTDinterwordspacing

\bibitem{Bloomfield_2012}
\BIBentryALTinterwordspacing
D.~S. Bloomfield \emph{et~al.}, ``Toward reliable benchmarking of solar flare
  forecasting methods,'' \emph{The Astrophysical Journal}, vol. 747, no.~2, p.
  L41, Feb 2012. [Online]. Available:
  \url{https://iopscience.iop.org/article/10.1088/2041-8205/747/2/L41}
\BIBentrySTDinterwordspacing

\bibitem{benvenuto2018hybrid}
\BIBentryALTinterwordspacing
F.~Benvenuto, M.~Piana, C.~Campi, and A.~M. Massone, ``A hybrid
  supervised/unsupervised machine learning approach to solar flare
  prediction,'' \emph{The Astrophysical Journal}, vol. 853, no.~1, p.~90, jan
  2018. [Online]. Available:
  \url{https://doi.org/10.3847\%2F1538-4357\%2Faaa23c}
\BIBentrySTDinterwordspacing

\bibitem{woodcock1976evaluation}
\BIBentryALTinterwordspacing
F.~Woodcock, ``The evaluation of yes/no forecasts for scientific and
  administrative purposes,'' \emph{Monthly Weather Review}, vol. 104, no.~10,
  pp. 1209--1214, 1976. [Online]. Available:
  \url{https://doi.org/10.1175/1520-0493(1976)104<1209:TEOYFF>2.0.CO;2}
\BIBentrySTDinterwordspacing

\bibitem{barnes2008evaluating}
\BIBentryALTinterwordspacing
G.~Barnes and K.~Leka, ``Evaluating the performance of solar flare forecasting
  methods,'' \emph{The Astrophysical Journal Letters}, vol. 688, no.~2, p.
  L107, 2008. [Online]. Available: \url{https://doi.org/10.1086/595550}
\BIBentrySTDinterwordspacing

\bibitem{mason2010testing}
\BIBentryALTinterwordspacing
J.~Mason and J.~Hoeksema, ``Testing automated solar flare forecasting with 13
  years of michelson doppler imager magnetograms,'' \emph{The Astrophysical
  Journal}, vol. 723, no.~1, p. 634, 2010. [Online]. Available:
  \url{https://doi.org/10.1088/0004-637X/723/1/634}
\BIBentrySTDinterwordspacing

\bibitem{ben2010user}
\BIBentryALTinterwordspacing
A.~Ben-Hur and J.~Weston, ``A user's guide to support vector machines,''
  \emph{Methods in molecular biology (Clifton, N.J.)}, vol. 609, pp. 223--39,
  01 2010. [Online]. Available:
  \url{doi.org/https:10.1007/978-1-60327-241-4_13}
\BIBentrySTDinterwordspacing

\end{thebibliography}

\end{document}